\pdfoutput=1
\documentclass[runningheads]{llncs}
\usepackage{graphicx}
\DeclareGraphicsExtensions{.pdf,.png,.jpg, .eps}
\usepackage{epstopdf}
\usepackage{hyperref}
\usepackage{amssymb}
\usepackage{amsmath}
\usepackage{url}
\usepackage[caption=false]{subfig}
\urldef{\mailsa}\path|martin.ringsquandl.ext|
\urldef{\mailsb}\path||
\urldef{\mailsc}\path|@siemens.com|    
\newcommand{\keywords}[1]{\par\addvspace\baselineskip
\noindent\keywordname\enspace\ignorespaces#1}
\normalsize
\begin{document}

\mainmatter  % start of an individual contribution

\title{Context-Aware Analytics in MOM Applications}

\titlerunning{Context-Aware Analytics in MOM Applications}

\author{Martin Ringsquandl
%
%\thanks{Please note that the LNCS Editorial assumes that all authors have used
%the western naming convention, with given names preceding surnames. This determines
%the structure of the names in the running heads and the author index.}%
\and Steffen Lamparter
\and Raffaello Lepratti}

\institute{Corporate Technology\\Siemens AG\\
Munich, Germany\\ \mailsa \mailsb \mailsc}

%Weaknesses: 
%- How to integrate into single ontology?
%- How does the ontology change?
%- How to track changes in history?
%- How to couple data with ontology?
% -> Clusters of context/situations might decrease complexity
%

\toctitle{Context-aware Analytics in MOM Applications}
\tocauthor{Authors' Instructions}
\maketitle
%%
%% Proceedings Production
%%
\setcounter{page}{44}
\thispagestyle{plain}

\hypersetup{
	pdfauthor = {Ringsquandl, Martin; Lamparter, Steffen; Lepratti, Raffaello},
	pdftitle = {Context-aware Analytics in MOM Applications},
	pdfkeywords = {Semantic Context, Semantic Integration, Context-aware Analytics, Manufacturing Operations Management)}
	}

\begin{abstract}  
	Manufacturing Operations Management (MOM) systems are complex in the sense that they integrate data from heterogeneous systems inside the automation pyramid. The need for context-aware analytics arises from the dynamics of these systems that influence data generation and hamper comparability of analytics, especially predictive models (e.g. predictive maintenance), where concept drift affects application of these models in the future. \\
	Recently, an increasing amount of research has been directed towards data integration using semantic context models.	Manual construction of such context models is an elaborate and error-prone task. Therefore, we pose the challenge to apply combinations of knowledge extraction techniques in the domain of analytics in MOM, which comprises the scope of data integration within Product Life-cycle Management (PLM), Enterprise Resource Planning (ERP), and Manufacturing Execution Systems (MES). We describe motivations, technological challenges and show benefits of context-aware analytics, which leverage from and regard the interconnectedness of semantic context data. Our example scenario shows the need for distribution and effective change tracking of context information.
\end{abstract}

%
% -------------------------Introduction----------------------------
%
\section{Introduction}
\label{intro}
Within data analytics on top of complex MOM applications, such as combining supply chain and automation data, IT systems need to integrate vast amounts of data which are generated by several different sources (e.g. RFID sensors, purchase orders) using many distinguished (un)structured data models. Analytics such as production forecasts require more sophisticated models, e.g. machine learning, that leverage from this interconnected data that is currently hidden in heterogeneous silo systems. Manufacturers are facing challenges, for example, in quality assurance and detection of non-compliance of incoming materials which are delivered by various suppliers. In such a scenario, they need to analyze feedback about defective products and production context at the corresponding plant to effectively employ predictive models. So far, such models do not account for context changes over time of the underlying systems between individual samples on both sides (supplier and manufacturer), e.g. exchange or repair of devices. 
\\
In general, as MOM systems produce lots of context information, the significance of analytic models that only incorporate parts of this data in isolation becomes more fragile.  
To overcome this lack of integrated operations and analytics, there is not only a need for data integration in business to business (B2B), i.e. along the supply chain, but also in business to manufacturing (B2M). One starting point for such integration is the set of models and terminologies defined in ANSI/ISA-95 Integration Standard and the resulting IEC 62264. These models were developed to facilitate interfaces and data integration of business applications into manufacturing systems \cite{IEC62264}. Additionally, with the emergence of new information model standards like OPC UA and AutomationML, MOM systems already are able to provide standardized meta-data descriptions. \\ In this paper, we pose the challenge for the extension and semantic lifting of existing MOM information models, particularly focusing on their value in the application of drifting analytic concepts. 
%Extension means to extract additional knowledge that is currently not included in standard information models. Semantic lifting aims to provide formally defined semantics and harmonization of metadata representations in existing information models. This facilitates interoperability, reusability and .
%Typically, quality control plans consist of:
%\begin{itemize} 
%	\item work verification (worker has appropriate qualifications), 
%	\item material receiving (verifying incoming material $\rightarrow$ outsourced to supplier),
%	\item supplier qualification (has to be qualified to ISO XX),
%	\item quality feedback (customer complaints$ \rightarrow$ non-compliance),
%	\item corrective action (why didn't this get captured by the quality assurance? $\rightarrow$ improve quality plan)
%\end{itemize}
%the analyst does not have knowledge about past situations
%Not only change of devices, but also change (maybe improvement) of processes.
%The same counts for material demand forecasts based on previous customer demand. If processes or machines change, the situations can not be compared, should be adapted to the context change (discount factor?). 

%
% -------------------------Related Work----------------------------
%
\section{Related Work}
\label{rel_work}
There are well-known machine learning approaches for the detection of changes, also called concept drift, in underlying data generation processes \cite{Dries2009}. Closely related to our problem formulation, the work of Kiseleva discusses the need for context-awareness in predictive models for user actions on websites \cite{Kiseleva2013}. However, these approaches are too restrictive in the case of MOM, because context information and its changing conditions could in fact be made available, so there is no need to repeatedly use statistical tests or clustering for \textit{hidden} concept drifts. Nevertheless, we want to consider these established approaches for verification purposes.

Since we identify the need for data integration based on a semantic lifting of existing data sources, another area of research that relates to our work is the usage of ontologies and semantic data integration. A concise overview of semantic integration can be found in \cite{Noy2004} and \cite{Cruz2005}. 
%
%---------------------Preliminaries--------------------------------
%
\section{Preliminaries}
\label{prelim}
\begin{figure}
%	\subfloat
%		\centering
%		\subfloat[a][a]{\includegraphics[height=5.5cm]{figures/mompyr.pdf}\label{fig:mompyr}}
%		\subfloat[b][b]{\includegraphics[height=3.5cm]{figures/final_picture.png}\label{<figure2>}}
%		\caption{Context of MOM systems in a classical automation pyramid}
%		\label{fig:both}
	\centering
	\includegraphics[height=5.5cm]{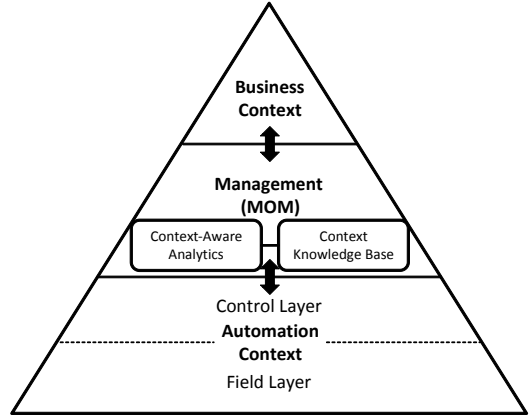}
	\caption{Context of MOM systems in a classical automation pyramid}
	\label{fig:mompyr}
\end{figure}

Our interpretation of context-aware analytics relates to Dey's definition of context, i.e. information that can be used to characterize the situation of an entity \cite{Dey2001}. Here, the entities are those which affect analytic models, more specifically, training data used by machine learning models. Figure \ref{fig:mompyr} shows that MOM systems are located in the middle of the classical automation pyramid of a manufacturing business. Therefore, they need to adapt to context information of the automation layers as well as the business layers. 
\paragraph{Context Knowledge Base}
In order to keep things as generic as possible, we define the context knowledge base $\mathcal{O}$ to be an ontology, e.g. it is based on some Description Logic (DL) language such as the OWL 2 DL base $\mathcal{SROIQ(D)}$. The concepts (TBox) of this context knowledge base could result from schema-level data integration setting over the different MOM data sources, see section \ref{cont_extr}. Dynamics of the underlying data sources are then reflected in the assertions (ABox) of the context knowledge base.
\paragraph{Context-aware Analytics}
Consequently, we again use a very generic definition of analytic models, however focusing on predictive machine learning models $M$, where output is used to make some decision about given input, $M : \mathcal{X} \rightarrow \mathcal{Y}$, 
with $\mathcal{X}$ and $\mathcal{Y}$ being the input and output space of the model, respectively. Such a model $M$ is trained on labeled source domain data set $D_s = \left\{ \langle \vec{x_1}, y_1 \rangle, \langle \vec{x_i}, y_i \rangle,... \langle \vec{x_n}, y_n \rangle \right\}$ with $\vec{x_i} \in \mathcal{X}, y_i \in \mathcal{Y}, i = 1,2...n$ and applied on data of an usually unknown destination domain.
\\
By treating input and output spaces as random variables, the problem of concept drift can be summarized as non-stationary probability distributions of source domain $P_s$ and destination domain $P_d$, as shown in \eqref{eq:cd}.
\begin{equation}
\label{eq:cd}
P_s(\mathcal{Y} \mid \mathcal{X}) \neq P_d(\mathcal{Y} \mid \mathcal{X}) \\
\end{equation}
\begin{claim}
	Given a sufficiently comprehensive context knowledge base $\mathcal{O}$, the conditional probability distribution $P(\mathcal{Y} \mid \mathcal{O})$ is stationary, i.e. $P_s(\mathcal{Y} \mid \mathcal{O}) = P_d(\mathcal{Y} \mid \mathcal{O})$.
\end{claim}
Following this notion, given a history of all context models $H = \left\{\mathcal{O}_1, \mathcal{O}_2, ..., \mathcal{O}_t \right\}$ tracked over time $t$ and a set of all possible analytic models $A = \left\{M_1, M_2, ..., M_k\right\}$ that can be employed for the same task, our goal is to find a mapping $\mathcal{F}$ from context to analytic model.
\begin{equation}
\label{eq:map}
\mathcal{F}: H \rightarrow A
\end{equation}
This function is necessary to find the best fitting analytic model for a given situation (context) and applies as soon as we want to classify an input vector $\vec{x'}$ with unknown output, sampled on underlying context model $\mathcal{O}_{t}$. Equations \eqref{eq:fun} show this situation, including the case when no such model can be found, i.e. search based on semantic similarity measures between contexts \cite{Oldakowski2005}.
\begin{equation}
	\label{eq:fun}
	\mathcal{F}(\mathcal{O}_{t}) = M^* \qquad
	\mathcal{F} (\arg\max_{\mathcal{O}_{t'} \in H} \mbox{sim}(\mathcal{O}_{t'}, \mathcal{O}_{t})) = M^*
\end{equation}
where $M^*$ is the optimal analytic model with respect to some performance criterion, $ \mathcal{O}_t$ is the current context and sim is a semantic similarity function.
%
% --------------------System Architecture-------------------------
%
\section{Example Scenario}
\label{exmp}
\begin{figure}
	\centering
	\includegraphics[height=4.8cm]{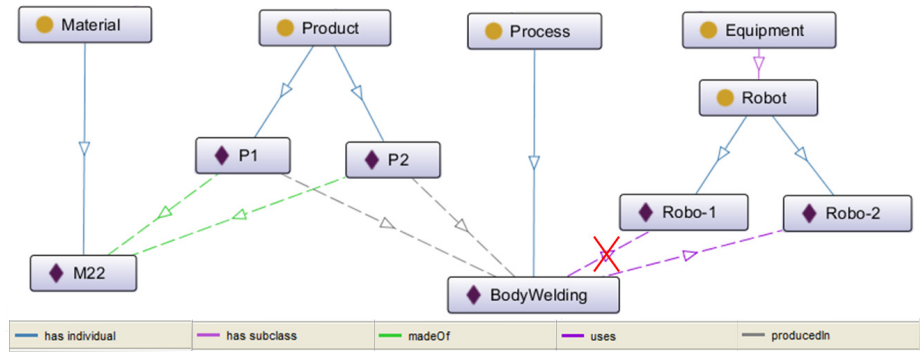}	
	\caption{Example Business-To-Manufacturing context plus evolving assertions}
	\label{fig:exmp}
\end{figure}

%Our proposed system architecture for context-aware analytics in MOM consists of a GaV context broker that integrates the schemas of business, automation and product life-cycle data sources. The context broker holds a knowledge base $\mathcal{O}$ that characterizes the current situation, a context history $H$ of previously extracted or apriori given context knowledge, and a concept drift verification module. 
As an example, consider a manufacturer of two different product types \textit{P1} and \textit{P2}, which both are made from supplied material aluminum \textit{M22}. The same welding robot \textit{Robo-1} is used for the construction (body welding) of both products. Figure \ref{fig:exmp} shows how this knowledge is asserted in a context knowledge base with concepts corresponding to ISA-95 standard. 
\\
As analytic application, the manufacturer wants to predict at arrival, whether incoming materials are going to cause a defective product, so predictive models are trained on historic quality assurance data like depicted in Table \ref{tab:data1} in order to learn the boolean concept of a defective product. Now, suppose the old robot is replaced at the shop floor by the new \textit{Robo-2} (hinted by the crossed role assertion in Figure \ref{fig:exmp}), which is much more tolerant to deviations in material and therefore produces only negative defect labels, i.e. the concept of defects is clearly drifting associated with context changing from $\mathcal{O}_t$ to $\mathcal{O}_{t'}$, as shown in Table \ref{tab:data2}. We denote this context change in the ABoxes $\mathcal{A}_t$ and $\mathcal{A}_{t'}$:
\begin{equation*}
\mathcal{A}_{t'} = (\mathcal{A}_t \setminus \left\{\mbox{uses(BodyWelding, Robo-1)}\right\}) \cup \left\{\mbox{uses(BodyWelding, Robo-2)}\right\}
\end{equation*}
\begin{table}
	\parbox{.45\linewidth}{
		\caption{Data before context change}
	\centering
	\begin{tabular}{ |c|c| }
		\hline
		Quality ($\vec{x}$) & Defect ($y$) \\
		\hline
		$\vec{x_1}$ & $true$ \\
		$\vec{x_2}$ & $false$ \\
		... & ... \\
		$\vec{x_n}$ & $true$ \\
		\hline 
	\end{tabular}
	$\begin{array}{l}
		\left.\rule{0pt}{6ex}\right\} \mathcal{O}_t
	\end{array}$
	
	\label{tab:data1}
	}
	\parbox{.45\linewidth}{
		\caption{Data after context change}
		\centering
		\begin{tabular}{ |c|c| }
			\hline
			Quality ($\vec{x}$) & Defect ($y$)  \\
			\hline
			$\vec{x_1}$ & $false$ \\
			$\vec{x_2}$ & $false$ \\
			... & ... \\
			$\vec{x_n}$ & $false$ \\
			\hline 
		\end{tabular}
	$\begin{array}{l}
		\left.\rule{0pt}{6ex}\right\} \mathcal{O}_{t'}
		\end{array}$
	
	\label{tab:data2}
	}
\end{table}
%Figure \ref{fig:architecture} illustrates the integration at a manufacturer's perspective, where a context broker could be extended to communicate with external context brokers along the supply chain. To make sure that $H$ only tracks significant context changes, we can resort to the well-established concept drift approaches in machine learning (e.g. hypothesis tests, clustering) to verify if a change in the underlying data distributions had a significant effect and needs to be tracked in $H$, i.e. if the change from $\mathcal{O}_t$ to $\mathcal{O}_{t'}$ actually impacts analytic performance and consequently means that $\mathcal{F}(\mathcal{O}_t) \neq \mathcal{F}(\mathcal{O}_{t'})$.
%
%-------------------Context Extraction-----------------
%
\section{Context Extraction}
\label{cont_extr}
We present an overview of information models used in MOM systems to support the claim that some context information can readily be made available. Semantic liftings of local MOM data sources seem to be promising, because they can easily be mapped to a global context knowledge base. This allows the analytic models to be aware of context changes in any of the underlying data sources. Examples of local context changes are: replacement of field devices (Automation), switching material suppliers (Business), modified production processes (Life-cycle).
\paragraph{Context of Automation} 
In today's automation systems, there exist information model languages for the definition of asset models, event hierarchies, and more, such as OPC UA. Vendors can ship devices with their own pre-configured information model, which can then be automatically interpreted by other devices in the plant that comply to the same standard. Field device descriptions of OPC UA could be used in our example scenario in section \ref{exmp}. Since a flexible data model is crucial to incorporate dynamic changes and due to their similarities, we see a mapping from OPC UA to the Resource Description Framework (RDF).
%, where RDF \textit{Resource} maps to \textit{Node} in OPC UA, \textit{URIs} map to \textit{NodeIds} and a RDF \textit{Statement} is mapped to \textit{Reference}. 
%A meta-model mapping like sketched in table \ref{tab:opc} allows the context broker to directly integration and distribute automation data.
%\begin{table}[t]
%\centering
%    \begin{tabular}{ | p{4cm} | p{4cm} |}
%    \hline
%    \textbf{OPC UA} & \textbf{RDF}\\ \hline
%    Node & Resource\\ \hline
%    NodeId & Unique Resource Identifier \\ \hline
%    Reference & Statement \\
%    \hline
%    \end{tabular}
%    \label{tab:opc}
%    \caption{Meta-model similarities between OPC UA and RDF}
%\end{table}
\paragraph{Context of Enterprise Applications}
Business context can be best reflected by inspecting ERP systems and their various data sources. For the purpose of this paper, it suffices to consider integration of context information from ERP and MES, like specified in ISA-95 standard (B2MML is an XML implementation of this standard).
There exist mappings from ERP systems data models to B2MML, plus this XML schema can be lifted to RDF or OWL \cite{Ferdinand2004}. 

\paragraph{Context of Life-cycle Management}
For PLM systems, a local meta-data representation can be made explicit by lifting the standard information models of AutomationML, which is based on the IEC62424 computer-aided engineering exchange (CAEX) format.
%
% ---------------------Context-aware Analytics------------------------
%
\section{Benefits}
\label{benefits}
%Having an explicit global context model Arguably any MOM application can benefit from the global context knowledge base, however, there still are challenges that need further investigation.
%\paragraph{Challenges}
%Considering the amount of possible context information available for every single data source, it becomes obvious that a context-aware analytic model cannot account for every context change. On the other hand, the significance of dynamics in the underlying data sources can be verified by concept drift approaches, which gradually reduces integration and historization efforts. Furthermore, the mapping from local schemata to the global context knowledge base is not clear.
Context-aware analytics regard situational dependencies, therefore they use more comparable, homogeneous data with less unexplained noise. Major benefits arise from the detection of recurring situations, where an analytic concept is drifting back and forth.
\begin{example}
	Consider again the scenario presented in section \ref{exmp}, where the defective product concept is drifting dependent on the welding robot. Assuming, model $M_1$ was used to learn the concept in context $\mathcal{O}_t$ and an updated model $M_2$ in context $\mathcal{O}_{t'}$. Suppose now that, due to maintenance of \textit{Robo-2}, the old robot \textit{Robo-1} is installed again and context changes back to $\mathcal{O}_t$. Since $\mathcal{F}(\mathcal{O}_{t}) = M_1$, we do not have to train a new model for this recurring concept and just switch to an already existing model.  
\end{example}  
Additionally, detection of concept drifts using well-established hypothesis tests can be employed to very if a context knowledge base is sufficiently comprehensive. More precisely, if an analytic concept is drifting without previously changing context, we know that the extracted context is incomplete and need to extend it. Therefore, the combination of concept drift and context knowledge is able to simultaneously define the scope of sufficiently comprehensive context and improve accuracy of analytics. 
%For the interaction between context knowledge and analytics, we see a high-level procedure that combines concept drift and context changes, as depicted in Figure \ref{fig:decision}. For each new instance to be classified in the current context, the analytic model has to be notified about context changes. If no changes happened, an existing (best matching) model can be employed. Otherwise, a concept drift verification should test for changes in the data distribution of the model's training data by comparing it to data sampled or bootstrapped after the context change happened. A verified concept drift results in a model adaption, i.e. re-training on post-concept drift data, and tracking of context changes. In case that the concept drift hypothesis can not be supported with enough evidence, the existing model can still be used. Additionally, if a concept drift is detected without previous changes in context, we know that the extracted context is incomplete and need to extend it.
%\begin{figure}
%\centering
%	\includegraphics[height=6cm]{figures/decision.pdf}	
%		\caption{Interaction between context and analytics}
%		\label{fig:decision}
%\end{figure}
%	\item data analysts do not need to fully understand and dive deep into the characteristics of data sets,
%	\item reduced data extraction, transformation, loading (ETL) efforts.
\section{Conclusion}
\label{concl} 
By motivating and defining context-awareness in data analytics, we discussed the value of context information for particular scenarios in MOM environments. The problem setting incorporates semantic data integration and a complementary use of machine learning and extraction of context knowledge. Following this problem setting, we argue that both analytic and context models can iteratively be optimized. Future work includes development of concrete algorithms, according implementation and evaluation.
\label{references}
\bibliographystyle{plain}
\bibliography{library}
\end{document}